\documentclass[smallcondensed]{svjour3}
\pdfoutput=1

\def\journal{0}   

\usepackage{amsmath,amssymb}
\usepackage{empheq}
\usepackage{booktabs}
\usepackage[english]{babel}
\usepackage{enumitem}
\usepackage[left=32mm, right=32mm, top=32mm, bottom=32mm]{geometry}
\usepackage{graphicx}
\usepackage[bookmarks=false]{hyperref}
\usepackage{lmodern}
\usepackage{caption, subcaption}
\usepackage{url}

\usepackage{natbib}
\usepackage{algorithm,algpseudocode}

\setlist{itemsep=0.5ex}

\numberwithin{equation}{section}

\newcommand{\etabold}{{\boldsymbol \eta}}
\newcommand{\chibold}{{\boldsymbol \chi}}
\newcommand{\alphabold}{{\boldsymbol \alpha}}
\newcommand{\thetabold}{{\boldsymbol \theta}}
\newcommand{\Thetabold}{{\boldsymbol \Theta}}
\newcommand{\T}{\mathbf{T}}
\newcommand{\x}{\mathbf x}
\newcommand{\X}{\mathbf{X}}
\newcommand{\R}{\mathbb{R}}
\newcommand{\diff}{\operatorname{d}\!}
\newcommand{\argmax}{\operatorname{arg\,max}\ }

\renewcommand{\S}[1]{$\text{Boj}_{#1}$}
\newcommand{\transpose}{{\rm T}}

\newcommand{\emails}{{kaspar.thommen@ \{ \href{mailto:kaspar.thommen@ubs.com}{ubs.com}, \href{mailto:kaspar.thommen@gmail.com}{gmail.com} \}}}

\if 1\journal
  \journalname{Methodology and Computing in Applied Probability}
\else
  \journalname{}
\fi

\begin{document}

\title{A Closed-Form Approximation to the Conjugate Prior of the Dirichlet and Beta Distributions}
\author{Kaspar Thommen}
\institute{K. Thommen \at UBS AG \\ \email{\emails}}
\if 0\journal
  \date{\today}
\fi

\maketitle

\begin{abstract}
\noindent
We derive the conjugate prior of the Dirichlet and beta distributions and explore it with numerical examples to gain an intuitive understanding of the distribution itself, its hyperparameters, and conditions concerning its convergence. Due to the prior's intractability, we proceed to define and analyze a closed-form approximation.
Finally, we provide an algorithm implementing this approximation that enables fully tractable Bayesian conjugate treatment of Dirichlet and beta likelihoods without the need for Monte Carlo simulations.
\keywords{Bayesian \and conjugate prior \and Dirichlet distribution \and beta distribution \and tractable \and closed-form \and MAP approximation}
\subclass{62F15 \and 65C60 \and 62E17}
\end{abstract}


\section{Introduction and prior work}
The probabilistic modeling of proportional or compositional data (i.e., vectors containing fractions that sum up to 1) arises in many fields: in finance, we might be interested in modeling the composition of client portfolios with respect to assets classes contained therein, and in political sciences or A/B testing, the probabilistic analysis of voting fractions may be of interest.

Fractional observations can often be modeled using Dirichlet (or beta) likelihoods. To enable computationally efficient Bayesian inference (e.g. for changeppoint detection, see \citep{adams2007bayesian}) a conjugate prior is desired. Recently, \citep{andreoli2018conjugate} has derived the prior of the Dirichlet distribution, which he has named \emph{Boojum} distribution. However, due to its intractability, the Boojum distribution prevents itself from many practical applications due to the necessary resource-intensive Monte Carlo simulations. We overcome this problem by providing a closed-form approximation that renders the Boojum-Dirichlet conjugate pair tractable.

\section{The conjugate prior of the Dirichlet distribution}
First, we recall the systematic construction procedure of conjugate priors for exponential family distributions. Then, we apply this general solution to the Dirichlet likelihood in order to find its conjugate prior, the Boojum distribution. Note that this derivation also holds for beta likelihoods by setting the dimensionality of the problem to $D = 2$.

\subsection{Conjugate priors of exponential family distributions}
Given a likelihood from the exponential family defined by
\begin{equation}
p_L(\x \mid \etabold) = h(\x) g(\etabold) \exp \left({\etabold}^\transpose \T(\x) \right) \label{eq:likelihood}
\end{equation}
for the vector random variable $\x \in \R^D$, with parameter vector $\etabold \in \R^D$ and known functions $h(\x): \R^D \mapsto \R$, $g(\etabold): \R^D \mapsto \R$ and $\T(\x): \R^D \mapsto \R^D$, there is a conjugate prior of the form
\begin{equation}
p_\pi(\etabold \mid \chibold, \nu) = f(\chibold, \nu) g(\etabold)^\nu \exp \left(\etabold^\transpose \chibold \right) \label{eq:prior}
\end{equation}
with $\chibold \in \R^D$ and $\nu \in \R > 0$ representing prior hyperparameters \citep{keener2010theoretical}. The normalizing constant is given by
\begin{equation}
f(\chibold, \nu) = \left( \int_\etabold g(\etabold)^\nu \exp \left(\etabold^\transpose \chibold \right) \diff \etabold \right)^{-1}.
\end{equation}
The posterior after having collected $n$ observations $\X = \{\x_i\}_{i=1}^n$ is, thanks to conjugacy,
\begin{equation}
p(\etabold \mid \X, \chibold, \nu) = p_\pi (\etabold \mid \chibold', \nu')
\end{equation}
with updated hyperparameters
\begin{align}
\chibold' &= \chibold + \sum_{i=1}^n \T(\x_i)\\
\nu' &= \nu + n.
\end{align}
The prior predictive distribution (or the posterior predictive distribution if appropriately updated hyperparameters $\nu'$ and $\chibold'$ are employed instead) is defined as follows:
\begin{equation}
p_\text{pr}(\x \mid \chibold, \nu)
=
\int_\etabold
p_\text{L}(\x \mid \etabold)
\cdot
p_\pi(\etabold \mid \chibold, \nu)
\diff\etabold.  \label{eq:predictive}
\end{equation}

\subsection{Application to the Dirichlet distribution}
The likelihood function of the Dirichlet distribution parameterized with $\alphabold \in \R^D$ for a random variable $\thetabold \in \R^D$ belonging to the $D-1$ simplex (i.e., $\sum_{k=1}^D \thetabold_{[k]} = 1$ with $\thetabold_{[k]}$ representing the $k$-th vector entry) is given by
\begin{align}
p_\text{L,Dir}(\thetabold \mid \alphabold)
  &= B(\alphabold)^{-1} \prod_{k=1}^D \thetabold_{[k]}^{\alphabold_{[k]}-1} \\
  &= B(\alphabold)^{-1} \exp \left(\sum_{k=1}^D \alphabold_{[k]} \log \thetabold_{[k]} \right) \prod_{k=1}^D \thetabold_{[k]}^{-1} \\
  &= B(\alphabold)^{-1} \exp \left(\alphabold^\transpose \log \thetabold \right) \prod_{k=1}^D \thetabold_{[k]}^{-1} \label{eq:likelihood_dir}
\end{align}
with $B(\alphabold)$ representing the multivariate beta function and where we have introduced the notation
\begin{equation}
\log \thetabold \equiv \left( \log \thetabold_{[1]}, \log \thetabold_{[2]}, \dots, \log \thetabold_{[D]} \right)^\transpose. \label{eq:vector_log}
\end{equation}
Matching terms of (\ref{eq:likelihood_dir}) with (\ref{eq:likelihood}) yields the following identities:
\begin{align}
\x &\equiv \thetabold \\
\etabold &\equiv \alphabold \\
h(\x) &= h(\thetabold) \equiv \prod_{k=1}^D \thetabold_{[k]}^{-1} \\
g(\etabold) &= g(\alphabold) \equiv B(\alphabold)^{-1} \\
\T(\x) &= \T(\thetabold) \equiv \log \thetabold .
\end{align}
This allows us to rewrite the generic prior (\ref{eq:prior}) for the Dirichlet distribution, which yields the definition of the Boojum distribution:
\begin{equation}
\boxed{
  p_{\text{Boj}}(\alphabold \mid \chibold, \nu) = f(\chibold, \nu) B(\alphabold)^{-\nu} \exp \left(\alphabold^\transpose \chibold \right) \label{eq:boojum}
}
\end{equation}
with
\begin{equation}
f(\chibold, \nu) = \left( \int_\alphabold B(\alphabold)^{-\nu} \exp \left(\alphabold^\transpose \chibold \right) \diff \alphabold \right)^{-1}. \label{eq:normalizing_constant}
\end{equation}
The posterior has the same functional form as (\ref{eq:boojum}) due to conjugacy, but operates on updated hyperparameters:
\begin{subequations}
\begin{empheq}[box=\fbox]{align}
\nu &\rightarrow \nu' = \nu + n \label{eq:nu_post} \\
\chibold &\rightarrow \chibold' = \chibold + \sum_{i=1}^n \log \thetabold_i \label{eq:chi_post}
\end{empheq}
\end{subequations}
Finally, the predictive distribution (\ref{eq:predictive}) translates to
\begin{equation}
p_\text{pr,Boj}(\thetabold \mid \chibold, \nu)
=
\int_\alphabold
p_\text{L,Dir}(\thetabold \mid \alphabold)
\cdot
p_\text{Boj}(\alphabold \mid \chibold, \nu)
\diff\alphabold. \label{eq:boojum_predictive}
\end{equation}

\subsection{Hyperparameter interpretation} \label{sec:hyperparameters}
We can interpret the prior hyperparameters $\nu$ and $\chibold$ as summary statistics of a set of $n_\pi$ prior pseudo-observations $\thetabold_{\pi,i}$ computed using the posterior update equations (\ref{eq:nu_post}) and (\ref{eq:chi_post}), see Table {\ref{tab:hyperparameters}}.

\begin{table}[!htb]
\centering
\begin{tabular}{l p{70mm} l}
\toprule
Hyperparameter  &  Interpretation / summary statistic                                                                & Value \\
\midrule
$\nu$           &  Number of prior pseudo-observations                                                               &  $n_\pi$ \\
$\chibold$      &  Sum of vector-logs (see (\ref{eq:vector_log})) of prior pseudo-observations $\thetabold_{\pi,i}$  &  $\displaystyle \sum_{i=1}^{n_\pi} \log \thetabold_{\pi,i}$  \\
\bottomrule
\end{tabular}
\caption{Boojum hyperparameters}
\label{tab:hyperparameters}
\end{table}

\noindent Generally, given an arbitrary set of observations
\begin{equation}
\Thetabold \equiv \{ \thetabold_i \}_{i=1}^n \label{eq:Thetabold}
\end{equation}
and after dropping the ``prior'' qualifier and the $(\cdot)_\pi$ subscript, we can compute both $\nu$ and $\chibold$ according to Table \ref{tab:hyperparameters}. In other words, \emph{an observation set $\Thetabold$ fully defines a Boojum distribution}. In case of a prior, $\Thetabold = \Thetabold_\pi$ is the set of prior pseudo-observations, and for Boojum posteriors the set $\Thetabold$ comprises both prior pseudo-observations and actual observations.

\subsection{Convergence analysis} \label{sec:convergence}
The integral defining the Boojum distribution's normalizing constant (\ref{eq:normalizing_constant}) does not converge for all hyperparameter values $\nu$ and $\chibold$ as demonstrated by \citep{andreoli2018conjugate}. He finds that all of the following conditions must hold:
\begin{itemize}
\item (a) $\chibold_{[k]} < 0 \quad \forall k \in \{ 1, 2, \dots, D \}$
\item (b) $\nu > -1$
\item (c) $\nu \leq 0$ \quad \emph{or} \quad $\sum_{k=1}^D \exp \left( \chibold_{[k]}/\nu \right) < 1 \quad \forall k \in \{ 1, 2, \dots, D \}$
\end{itemize}
If we define the Boojum hyperparameters by means of prior pseudo-observations as proposed in Section  \ref{sec:hyperparameters}, we find that:
\begin{itemize}
\item (a) is satisfied: $\thetabold_{i [k]} < 1 \Rightarrow \log \thetabold_{i [k]} < 0 \Rightarrow \chibold_{[k]} = \sum_{i=1}^n \log \thetabold_{i [k]} < 0$
\item (b) is satisfied: $\nu = n > 0 > -1$
\end{itemize}
Evaluating condition (c) is more involved. The left-hand side sub-condition, $\nu \leq 0$, is clearly violated given that $\nu = n > 0$, so we must evaluate the right-hand side:
\begin{align}
\sum_{k=1}^D \exp \left( \chibold_{[k]}/\nu \right)
&= \sum_{k=1}^D \exp \left( \chibold_{[k]}/n \right) \\
&= \sum_{k=1}^D \left( \exp \left( \chibold_{[k]} \right) \right) ^{1/n} \\
&= \sum_{k=1}^D \left( \exp \left( \sum_{i=1}^n \log \thetabold_{i [k]} \right) \right)^{1/n} \\
&= \sum_{k=1}^D \left( \prod_{i=1}^n \thetabold_{i [k]} \right)^{1/n} \label{eq:condition_c}
\end{align}
Note that the summands are the \emph{geometric means} of the $k$-th entries across the $n$ observation vectors used to define the Boojum distribution. In other words, the right-hand side of (\ref{eq:condition_c}) equals the sum over the components of the ``geometric mean vector'' of the prior pseudo-observations.

For any set of positive numbers, the geometric mean is less than or equal to the arithmetic mean, hence we can write (\ref{eq:condition_c}) as
\begin{align}
\sum_{k=1}^D \exp \left( \chibold_{[k]}/\nu \right)
&=    \sum_{k=1}^D \left( \prod_{i=1}^n \thetabold_{i [k]} \right)^{1/n} \\
&\leq \sum_{k=1}^D \left(\frac{1}{n} \sum_{i=1}^n \thetabold_{i [k]} \right) \label{eq:condition_c2} \\
&=    \frac{1}{n} \sum_{i=1}^n \sum_{k=1}^D \thetabold_{i [k]} \\
&=    \frac{1}{n} \sum_{i=1}^n 1 \\
&=    1.
\end{align}
Because the geometric mean of a set of positive numbers only equals the arithmetic mean if the set is composed of identical numbers, we can conclude the following:
\begin{itemize}
\item If $n = 1$ or if $n > 1$ and all $\thetabold_i$ are identical, the geometric and arithmetic means are equal, thus rendering (\ref{eq:condition_c2}) an equality. Hence, in this case, condition (c) is \emph{not satisfied}.
\item Otherwise, i.e., if $n > 1$ and if not all observations are identical (i.e., when $\thetabold_{i [k]} \neq \thetabold_{j [k]}$ for any $i \neq j$ and any $k$), then the arithmetic mean dominates its geometric counterpart. Hence, in this case, the left-hand side of (\ref{eq:condition_c2}) in  is strictly less than the right-hand side, and it follows that condition (c) is \emph{satisfied}.
\end{itemize}
To summarize, the Boojum distribution only converges if we define it using \emph{two or more prior pseudo-observations that are not all identical}. Section \ref{sec:analysis} will present visualizations of the Boojum distribution that demonstrate this finding graphically.

\subsection{MAP approximation} \label{sec:map_approximation}
The Boojum's normalizing constant (\ref{eq:normalizing_constant}) cannot be computed analytically. This leads to the intractability of the distribution itself and derived distributions such as the predictive distribution (\ref{eq:boojum_predictive}). In order to avoid inefficient Monte Carlo simulations, we seek to find a closed-form approximation to the Boojum distribution. We propose the \emph{maximum a posteriori} (MAP) method \citep{Murphy2012} which approximates the posterior probability distribution function (PDF) by a Dirac delta function located at its mode. After dropping the normalizing constant in (\ref{eq:boojum}), we can write the Boojum's mode as
\begin{align}
\alphabold_\text{MAP}(\nu, \chibold)
 &= \underset{\alphabold}{\argmax}      \left( B(\alphabold)^{-\nu} \exp \left(\alphabold^\transpose \chibold \right) \right) \\
 &= \underset{\alphabold}{\argmax} \log \left( B(\alphabold)^{-\nu} \exp \left(\alphabold^\transpose \chibold \right) \right) \\
 &= \underset{\alphabold}{\argmax} \left(\alphabold^\transpose \chibold -\nu \log B(\alphabold) \right) \\
 &= \underset{\alphabold}{\argmax} \left(\alphabold^\transpose\frac{\chibold}{\nu} - \log B(\alphabold) \right) \label{eq:alpha_map}
\end{align}
which shows that $\alphabold_\text{MAP}$ is effectively a function of a single (vector) variable only, namely $\chibold / \nu$. Setting the derivative with respect to $\alphabold_{[k]}$ of the argument in (\ref{eq:alpha_map}) to zero yields
\begin{align}
0 &= \frac{\partial}{\partial \alphabold_{[k]}} \left( \alphabold^\transpose \frac{\chibold}{\nu} -\log B(\alphabold) \right) \\
  &= \frac{\chibold_{[k]}}{\nu} - \frac{1}{B(\alphabold)} \frac{\partial}{\partial \alphabold_{[k]}} B(\alphabold) \\
  &= \frac{\chibold_{[k]}}{\nu} - \frac{1}{B(\alphabold)} \left[ B(\alphabold) \left( \psi(\alphabold_{[k]}) - \psi \left( \sum_{\ell=1}^D \alphabold[\ell] \right) \right) \right]  \\
  &= \frac{\chibold_{[k]}}{\nu} + \psi \left( \sum_{\ell=1}^D \alphabold[\ell] \right) -\psi(\alphabold_{[k]}) \;\; \forall k \in \{1, 2, \dots, D \} \label{eq:derivative}
\end{align}
where $\psi$ is the digamma function\footnote{$\displaystyle \psi(x) \equiv \frac{\diff}{\diff x} \log \Gamma(x)$ where $\Gamma(x)$ is the gamma function.}. This set of $D$ dependent equations lacks an analytic solution for $\alphabold$, so we must resort to numerical methods. To this end, either gradient ascent methods operating on the derivative (\ref{eq:derivative}) (e.g. the \emph{Adam} optimizer \citep{kingma2017adam} which is popular for its good performance in the neural network realm) or direct optimization methods (e.g. the \emph{Nelder-Mead} method \citep{NelderMead65}) can be employed.
\enlargethispage{\baselineskip}

Once $\alphabold_\text{MAP}$ is determined, the MAP approximation of the Boojum posterior becomes, by definition,
\begin{equation}
p_\text{Boj,MAP}(\alphabold \mid \chibold, \nu) = \delta(\alphabold_\text{MAP}(\nu, \chibold)) \label{eq:boojum_approx}
\end{equation}
where $\delta(\cdot)$ is the Dirac delta function located at its argument.
Consequently, the posterior predictive distribution (\ref{eq:boojum_predictive}) simplifies to the likelihood evaluated at the mode $\alphabold_\text{MAP}$, thus rendering it tractable (a welcome side-effect of the MAP approximation):
\begin{equation}
p_\text{pr,Boj,MAP}(\thetabold \mid \chibold, \nu) = p_\text{L,Dir}(\thetabold \mid \alphabold_\text{MAP}(\nu, \chibold))  \label{eq:boojum_predictive_approx}
\end{equation}

This approximation will allow us to perform Bayesian inference of a Dirichlet or beta likelihood in closed-form (except for the optimization step in (\ref{eq:alpha_map}) that has to be carried out numerically). The next section will analyze the accuracy of the proposed MAP approximation.

\section{Analysis of the Boojum and the MAP approximation} \label{sec:analysis}
\subsection{Example scenarios}
We visualize the Boojum distribution as well the corresponding predictive distribution with numerical examples. This will help gaining an intuitive understanding of the Boojum distribution and the conditions affecting its convergence properties (see Section \ref{sec:convergence}). In order to be able to display the Boojum and all derived distributions graphically, we choose $D = 2$ dimensions for all examples, which simplifies the Dirichlet to a beta distribution.

Due to conjugacy, the Boojum PDF can be interpreted either as a prior distribution (with the observations set $\Thetabold$ representing prior pseudo-observations) or as a posterior distribution (where $\Thetabold$ is a combination of both prior pseudo-observations and actual observations). However, the MAP approximation to both the Boojum and and the corresponding predictive distribution necessitates the posterior interpretation by definition.

\begin{figure}[!htb]
  \begin{center}
  \includegraphics[width=0.9 \textwidth]{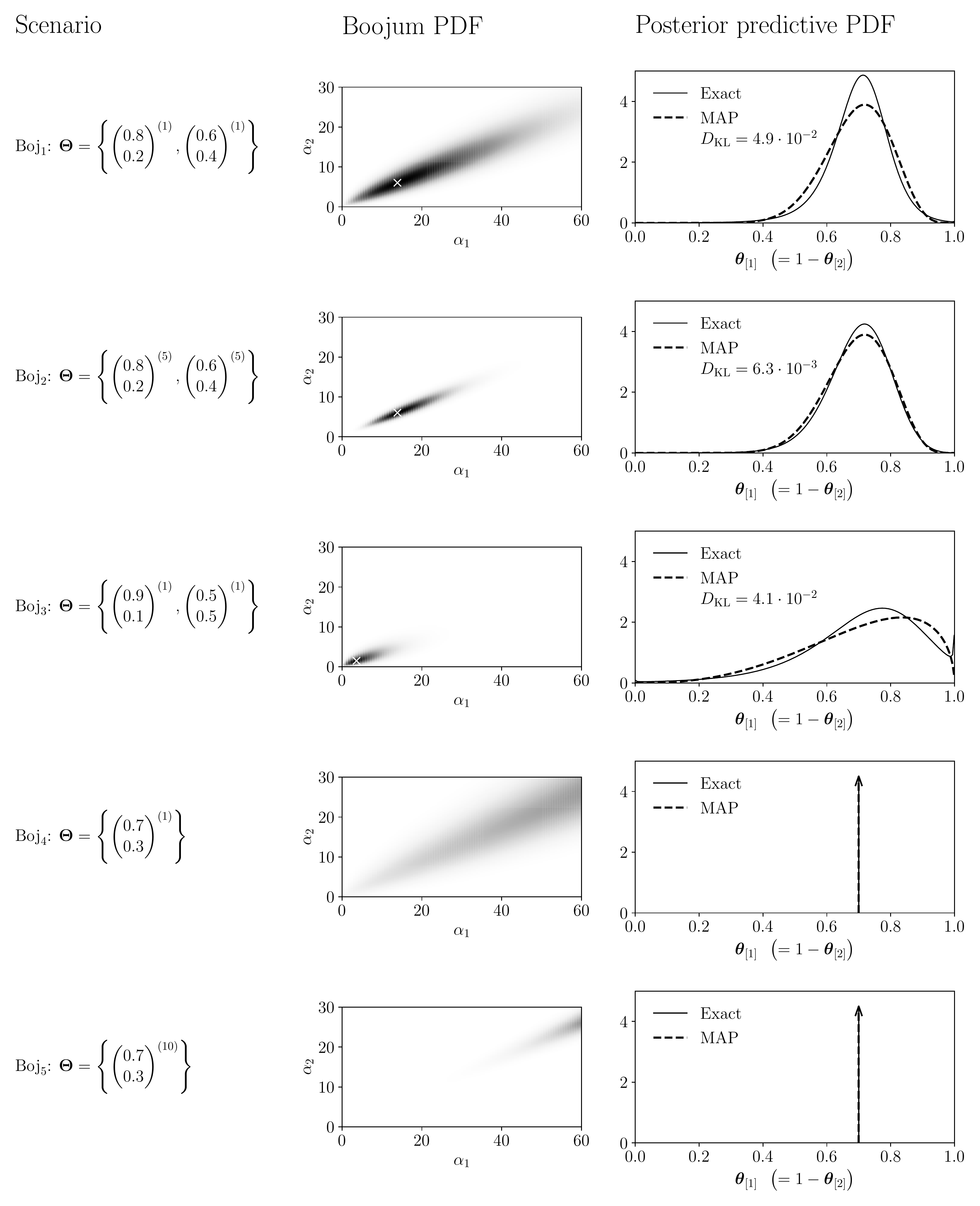}
  \caption[dummy]{Boojum scenarios configured with different observation sets $\Thetabold$ (see Section \ref{sec:hyperparameters}) and the resulting Boojum PDFs and posterior predictive PDFs respectively. Notes:
  \begin{itemize}
  \item Superscripts on the vectors in the definitions of $\Thetabold$ represent the multiplicity of the respective vector in the set.
  \item All scenarios have the same axes scaling in order to facilitate comparisons.
  \item The Boojum PDF has been normalized numerically. Darker areas indicate higher probability density. The maximum (mode) is calibrated to be black for \S{1}, \S{2} and \S{3}, which implies that the gray scales differ across these scenarios and thus cannot be directly compared. The gray scales for the improper scenarios \S{4} and \S{5} are arbitrary because the PDFs diverge.
  \item The white ``$\times$'' marks the mode of the Boojum PDF for the proper distributions \S{1}, \S{2} and \S{3}.
  \item The posterior predictive PDFs marked ``Exact'' have been obtained by numerical marginalization according to (\ref{eq:boojum_predictive}).
  \item We express the mismatch between the exact posterior predictive and the MAP using the Kullback-Leibler divergence, $D_\text{KL}$, as indicated in the plots. Note that the Kullback-Leibler divergence is not computable for the diverging scenarios \S{4} and \S{5}.
  \end{itemize} }
  \label{fig:scenarios}
  \end{center}
\end{figure}

Figure \ref{fig:scenarios} shows five Boojum distributions \S{1} through \S{5} configured with different sets of observations $\Thetabold$ that define the hyperparameters $\nu$ and $\chibold$ as per (\ref{eq:Thetabold}). We can make the following observations:
\begin{itemize}

\item \textbf{\S{1}:} The first scenario shows a Boojum parameterization with two distinct observations. The resulting Boojum PDF has the bulk of its probability density located approximately in the average direction of the two observation vectors, $(0.7, 0.3)^\transpose$. This somewhat surprising observation can be explained as follows: the parameterization of a Boojum prior (or any prior, for that matter) with a set of prior pseudo-observations naturally assigns these observations relatively high probabilities, simply because it has been defined by them. For $\alphabold_{[k]} > 1$, Dirichlet distributions have their probability density concentrated in the vicinity of the distribution's mean, $\alphabold / \lVert \alphabold \rVert$, hence $\alphabold$ must point in similar directions as the prior pseudo-observations.

Given that the probability density is quite spread out in the $\alphabold$ plane, the MAP approximation to the exact posterior predictive is rather coarse. This is clearly visible in the plot and can be quantified using the the Kullback-Leibler divergence $D_\text{KL}$ \citep{Kullback51klDivergence} which is also shown in the chart. The wide PDF implies thick tails in the exactly marginalized posterior predictive distribution, a feature that the MAP approximation doesn't exhibit.\footnote{Unfortunately, this fact is not very well visible in the plots due to their limited size.}

\item \textbf{\S{2}:} This scenario is based on the same observations that define \S{1}, but contains each of them five times respectively. As before, the resulting Boojum probability density is concentrated in the same average direction as the observations it is based on, but the higher number of observations leads to a more concentrated distribution. Note that the mode of \S{2} is identical to the mode of \S{1} because a simple increase in multiplicity does not affect the ratio $\chibold/\nu$ as per Table \ref{tab:hyperparameters}, which is effectively the dependent variable in the definition of $\alphabold_\text{MAP}$ (\ref{eq:alpha_map}).

The higher probability density concentration in \S{2} compared to \S{1} naturally puts the MAP approximation closer to the true posterior predictive as shown in the chart and the smaller Kullback-Leibler divergence.

\item \textbf{\S{3}:} Here we return to employing only two observations to define the Boojum distribution, but, unlike scenario \S{1}, the observations are spread further apart. Unsurprisingly, this wider variation leads to the probability density concentrated around shorter $\alphabold$ vectors that imply less peaked Dirichlet PDFs and that therefore assign more posterior predictive probability to a wider range of $\thetabold$ vectors, thus replicating the high variability of the observations used for configuring the Boojum in the first place.

Similar to \S{1}, we only see a moderately accurate fit between the exact posterior predictive distribution and the MAP approximation, again a consequence of the small number of observations that we have used to define the Boojum that has led to low concentration of probability density.

\item \textbf{\S{4}:} This case demonstrates a violation of convergence condition (c) in Section \ref{sec:convergence} by defining a Boojum distribution using only a single observations. Intuitively, this setup fails to ``teach'' the Boojum an appropriate measure of variability (or rather, we have indicated a desire for zero variability around the supplied observation). Indeed, the resulting Boojum PDF is improper: the probability density's peak diverges towards infinity (in the direction of the supplied observation vector), implying a preference for infinite $\alpha$.

Consequently, both the posterior predictive and its MAP approximation converge to a Dirac delta function located at the observation vector used in the definition of the Boojum distribution, $(0.7, 0.3)^\transpose$. This reflects our failure to configure the Boojum with a non-zero expected variability of observations.

\item \textbf{\S{5}:} The last scenario defines the Boojum using the same observation as \S{4} but repeats it ten times. As before, convergence condition (c) is violated, leading to similar conclusions as for \S{4}, but with faster divergence towards infinity caused by the greater number of observations.
\end{itemize}

\subsection{Summary of findings}
The numerical case studies analyzed in the previous section lead to the following conclusions for the construction of Boojum distributions using pseudo-observations:
\begin{itemize}
\item In order to satisfy all convergence criteria of Section \ref{sec:convergence}, Boojum priors must be constructed with at least two distinct prior pseudo-observations.
\item Encoding prior information about observation variability can be done by choosing prototypical prior pseudo-observation vectors with the desired variability. Note that this conclusion is common to \emph{all} conjugate priors irrespective of the likelihood, but recalling it helps to build an intuition around the rather novel Boojum distribution.
\item The proposed MAP approximation (\ref{eq:boojum_approx}) becomes more accurate the larger the number of observations encoded in the distribution (either through prior pseudo-observations or through actual observations).
\end{itemize}

\section{Algorithm}
We present 
an algorithm that performs closed-form\footnote{With the exception of the numerical computation of $\alphabold_\text{MAP}$ on line \ref{alg:alpha_map}.}, approximate Bayesian inference of Dirichlet and beta likelihoods using the MAP approximation of the Boojum prior.
\begin{algorithm}[!htb]
\caption{Approximate Bayesian conjugate inference for Dirichlet and beta likelihoods}
\begin{algorithmic}[1]
  \Require A set of $n_\pi$ prior pseudo-observations $\Thetabold_\pi = \{\thetabold_{\pi,1}, \thetabold_{\pi,2}, \dots, \thetabold_{\pi,n_\pi}\}$

  \State $\nu \gets n_\pi$
  \State $\chibold \gets \sum_{i=1}^{n_\pi} \log \thetabold_{\pi,i}$ \Comment{See (\ref{eq:vector_log})}

  \While{new observations $\thetabold_i$ arrive}
    \State $\nu \gets \nu + 1$ \Comment{See (\ref{eq:nu_post}) for $n = 1$}
    \State $\chibold \gets \chibold + \log \thetabold_i$ \Comment{See (\ref{eq:chi_post}) for $n = 1$ and (\ref{eq:vector_log})}
    \State $\alphabold_\text{MAP} \gets \underset{\alphabold}{\argmax} \left( B(\alphabold)^{-\nu} \exp \left(\alphabold^\transpose \chibold' \right) \right)$ \Comment{See Section \ref{sec:map_approximation}} \label{alg:alpha_map}
    \State \textbf{emit} $\alphabold_\text{MAP}$ \Comment{Emit posterior MAP estimate if required}
    \State \textbf{emit} $p_\text{pr,Boj,MAP}(\thetabold \mid \chibold, \nu)$ \Comment{Emit posterior predictive if required, see (\ref{eq:boojum_predictive_approx})}
  \EndWhile
\end{algorithmic}
\end{algorithm}

\noindent We supply a reference implementation of the above algorithm in the GitHub repository \citep{repo}.

\section{Conclusion and outlook}
We have derived a closed-form approximation to the Boojum distribution (i.e, to the the conjugate prior of the Dirichlet and beta likelihoods), including an exploratory analysis of the distribution and an algorithm to implement the procedure.

Further research should be directed at improving the MAP approximation, e.g. through variational inference, in an attempt to find more accurate posterior approximations that render it (and, ideally, the posterior predictive distribution) tractable.

\section{Data availability statement}
N/A

\begin{acknowledgements}
I want to thank my line manager Giuseppe Nuti for having given me the opportunity to work on this problem and for feedback on drafts of this paper. I also want to thank my colleagues Peter Larkin, Llu\'{i}s Jim\'{e}nez-Rugama and Mathias Brucherseifer for valuable feedback.
\end{acknowledgements}

\bibliography{bibliography}  
\end{document}